\documentclass[a4paper, 10pt, conference]{ieeeconf}      

\IEEEoverridecommandlockouts                              

\usepackage{amsmath} 
\newcommand{\RomanNumeralCaps}[1]
    {\MakeUppercase{\romannumeral #1}}
    
\usepackage[
  colorlinks,
  breaklinks,
  pdftitle={Eatron - Eatron},
  pdfauthor={Eatron},
  unicode
]{hyperref}
\usepackage[linesnumbered,ruled,vlined]{algorithm2e}

\SetCommentSty{mycommfont}

\SetKwInput{KwInput}{Input}                
\SetKwInput{KwOutput}{Output}              
\usepackage[noend]{algpseudocode}
\usepackage{graphicx}
\usepackage{float}
\usepackage{siunitx}

\title{\LARGE \bf Automated Lane Change Decision Making using Deep Reinforcement Learning in Dynamic and Uncertain Highway Environment}

\author{Ali Alizadeh$^{1^\ast}$, Majid Moghadam$^{2^\ast}$, Yunus Bicer$^{3^\ast}$, Nazim Kemal Ure$^{4}$, Ugur Yavas$^{5}$ and Can Kurtulus$^{5}$
\thanks{$^\ast$ These authors contributed equally to this work}
\thanks{$^{1}$A. Alizadeh is with Faculty of Mechatronics Engineering, Istanbul Technical University, Turkey {\tt\small Alizadeha at itu.edu.tr}}%
\thanks{$^{2}$M. Moghadam is with Department of Computer Engineering, University of California, Santa Cruz, USA {\tt\small mamoghad at ucsc.edu}}%
\thanks{$^{3}$Y. Bicer is with Faculty of Aeronautics and Astronautics, Aerospace Engineering, Istanbul Technical University, Turkey
        {\tt\small biceryu at itu.edu.tr}}%
\thanks{$^{4}$N.K. Ure is with ITU Artificial Intelligence and Data Science Research Center and Department of Aeronautical Engineering, Istanbul Technical University, Turkey
        {\tt\small ure at itu.edu.tr}}%
\thanks{$^{5}$U. Yavas and C. Kurtulus are with Eatron Technologies, Istanbul, Turkey {\tt\small Ugur.yavas, can.kurtulus at eatron.com}}
\thanks{   }
\thanks{IEEE Intelligent Transportation Systems Conference - ITSC 2019}
}

\begin{document}

\maketitle
\thispagestyle{empty}
\pagestyle{empty}

\begin{abstract}

Autonomous lane changing is a critical feature for advanced autonomous driving systems, that involves several challenges such as uncertainty in other driver's behaviors and the trade-off between safety and agility. In this work, we develop a novel simulation environment that emulates these challenges and train a deep reinforcement learning agent that yields consistent performance in a variety of dynamic and uncertain traffic scenarios. Results show that the proposed data-driven approach performs significantly better in noisy environments compared to methods that rely solely on heuristics.

\end{abstract}

\section{Introduction} \label{section:introduction}
Advanced Driving Assistance Systems (ADAS) are developed to increase traffic safety by reducing the impact of human errors. The evolution of various levels of driving autonomy has seen a significant speedup in last years aiming to enhance comfort, safety, and driving experience. For a long time, with a limited amount of technological resources, automotive stakeholders were focusing on steady-state maneuvers to achieve driving autonomy. However, in recent years one of the focuses of research in the field of autonomous driving is being directed to the transition maneuvers where tactical lane changing is an example, required for both fully and partially autonomous driving systems.

Many works have considered the automated lane changing problem as operational decision-making using control approaches ranging from vision-based to fuzzy and predictive control algorithms \cite{taylor1999comparative}, \cite{hatipoglu2003automated}, \cite{falcone2007predictive}. These methods approach the lane changing problem from a local perspective, which will overlook the highway traffic. For highway traffic, a tactical to strategic decision-making is necessary to capture the dynamics of the complex traffic environment such as reaching a goal distance like an exit, making lane change maneuver to avoid long-term traffic congestion while maintaining the safety criteria.

Autonomous lane change problem is mainly addressed by two different methods, rule-based and machine learning (ML). Rule-based methods were based on some predefined parameters that would tune the algorithm for a specific environment. On the other hand, ML methods have the capability of dealing with unforeseen situations after being well trained on a sufficient set of informative data. Since this work is using the learning methods, the surveyed literature will be narrowed to the ML-based approaches. Various learning approaches from end-to-end imitation learning \cite{bojarski2016end, bicer2019vision, Bicer2019sample} to Deep reinforcement learning (deep RL) \cite{sallab2017deep} have been applied to autonomous vehicles. Deep RL is efficient in learning arbitrary policies defining specific goals. In \cite{mukadamtactical}, a tactical decision making for lane changing in highway driving scenarios is being performed using deep RL. The agent is used to receive the occupancy grid of the entire simulation environment and produce the right/left lane change and accelerate/decelerate actions. The agent's performance seems promising compared with a simple rule-based approach and a human driver. However, the environment without vehicle dynamics seems oversimplified to train RL, where the observation space overlaps the entire simulation environment. This simplification could be the reason that they achieved fair results with a simple reward function.

DRL is also used in \cite{hoel2018automated} to automate the speed and lane change decision making, in which two different agents with different neural network architectures, 1-dimensional Convolution Neural Network (CNN) and Fully Connected Neural Network (FCNN), are developed and evaluated against the Intelligent Driving Model (IDM) \cite{treiber2000congested} and Minimize Overall Braking Induced by Lane changes (MOBIL) \cite{kesting2007general} algorithms. The agent receives observations that include relative position and velocities of all actors concerning the ego vehicle and produces the acceleration and right/left lane change actions. Unlike \cite{mukadamtactical}, the highway traffic environment used in this paper involves the dynamics of the cars and a low-level lateral controller for lane changing. The reward function is well-devised so that it will result in a comfortable and safe behaving agent. However, simple traffic tasks in which the actors drive in a single lane throughout the scenarios seem far from the realistic conditions. Besides, the ideal deterministic setting where observation space lacks uncertainties may deteriorate the validity of the agent's performance in more practical terms.

\subsection{Our contribution} \label{subsection:contribution}
Our contribution focuses on developing a deep RL agent that can robustly make safe lane changes in a dynamic highway driving setup while minimizing the estimated time of arrival (ETA). To this end, we set up a realistic environment in which the RL agent will be trained against a vast number of different scenarios, we incorporated an environment where each agent encompasses the dynamics of the vehicle, a reliable lateral controller, adaptive cruise controller (IDM \cite{treiber2000congested}), and a safe lane change algorithm called MOBIL \cite{kesting2007general}. The simulation environment is configurable with different uncertainty settings so that a realistic behavior can be simulated and analyzed.
In this context, the contribution of our paper is defined as follows:

\begin{itemize}
    \item Simulation results show that the performance of MOBIL is sensitive to the measurement noise and uncertainty in the environment. The proposed deep RL approach learns from traffic data and can yield robust performance in a large variety of uncertain and noisy traffic scenarios. 
    \item The developed simulation environment is capable of generating traffic scenarios where multiple cars execute lane and speed change decisions, while the agent receives noisy observations. Hence the proposed approach is more suitable for generating realistic traffic scenarios and training RL agents for autonomous driving, compared to the existing simulation settings that lack rich behavior in surrounding vehicles.   
\end{itemize}

\section{Background}

\subsection{Deep Q-Learning with real-time validation} \label{subsection:DQN}

Reinforcement Learning is a model-free algorithm to control an agent or a process to achieve the desired goal while interacting in a stochastic environment. The agent receives the observing states and rewards from the environment and tries to maximize the accumulated long-term return over the track of the interaction with environment \cite{busoniu2017reinforcement}. The backend of the RL is the Markov Decision Process (MDP) \cite{busoniu2017reinforcement}, which provides the mathematical framework to formalize the discrete stochastic environment.

Recent advances in Q-learning \cite{van2016deep, schaul2015prioritized, wang2015dueling, hessel2018rainbow} which is a model-free off-policy RL algorithm and the discrete nature of the action space along with the promising performance on planning \cite{moghadam2019hierarchical} and the control of autonomous systems \cite{lillicrap2015continuous} \cite{ure2019enhancing} motivated us to apply the deep version \cite{mnih2015human} of this algorithm to our problem.

Q-learning evaluates how good taking action might be at a particular state through learning the action-value function $Q(s, a)$. In Q-learning, a memory table $Q[s, a]$ is built to store the Q-values for all the possible combinations of states and actions. By taking action on the current state, the reward $R$ and the new states are acquired to take the next action that has the maximum $Q(s', a')$ in the memory table. Taking an action in a particular state will give a reward $R$ which is depicted in Eq. \ref{eq:target}.
\begin{equation}
\label{eq:target}
target = R(s, a, s') + \gamma \underset{a'}{\text{max}} Q_k (s', a')
\end{equation}
Where $s'$ and $a'$ are next state and action respectively. 
However, if the combinations of state and actions are too large or states and actions are continuous, the memory and computation requirement for action-value function $Q$ will be too high. To address this issue, Deep Q-Network (DQN) \cite{mnih2015human} is utilized that approximates the action-value function $Q(s, a)$.

In this work, we use the same algorithm introduced in DQN to train a neural network using the mini-batch samples from the experience replay. We also appended the real-time validation phase to the original DQN algorithm to record the best-trained model during the training. For this purpose, we define two periods, by which validation phase flag is activated, and the latest network weights are being recorded and exploited during validation. Depending on which period, the agent's performance is evaluated for several episodes, and the achieved mean reward is compared to the latest maximum value. This process enables the agent to record the best-trained model by validating on unseen scenarios. Defining two various periods with a different number of episodes helps the training to be faster and record a more generalized model at the same time. 

\section{Simulation Setup}
\begin{figure*}[h]
    \centering
    \includegraphics[width=2\columnwidth]{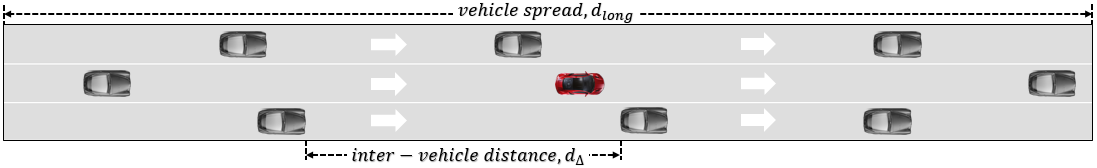}
    \caption{Highway initialization with inter-vehicle distance}
    \label{fig:Highway}
\end{figure*}

In this section, the devised dynamic traffic environment is described. Initialization logic, details of the kinematic model for the vehicle, two-point steering model, IDM and MOBIL algorithms are explained in detail.

\subsection{Initialization}
A highway with $n$ number of lanes is used as a simulation platform to develop our reinforcement learning agents. $m$ number of initial points are chosen randomly within maximum initial vehicle spread distance, $d_{long}$,  to determine the initial longitudinal ($x_0$) and lateral positions ($y_0$) of each vehicle with a minimum inter-vehicle distance, $d_{\triangle}$, as shown in Fig. \ref{fig:Highway}. Each vehicle has a dimension of $4.5 \times 2.5$ meter. Initial heading angle $\psi_0$ is chosen as 0 for all vehicles at the beginning. Vehicles are ordered by their longitudinal positions, and the middle vehicle is chosen as the ego vehicle. For the vehicles behind the ego vehicle, random initial speeds, $v_0$ are defined within the range of $[v^{rear}_{min},v^{rear}_{max}]$, same logic also applied for the vehicles in front and ego vehicle itself with the ranges of $[v^{front}_{min},v^{front}_{max}]$ and $[v^{ego}_{min},v^{ego}_{max}]$. At the last part of the initialization, desired speed are defined for each vehicle within the ranges of $[v^{d}_{min},v^{d}_{max}]$ and $v^{d}_{ego}$. Each episode has a length of $d_{max}$. These parameters are mostly taken from \cite{hoel2018automated}. The parameter values are explicitly depicted in Table \ref{table:Highway}. 

\begin{table}[htb]
\caption{Highway Simulation Parameters} 
\vspace*{-2mm}
\centering    
\begin{tabular}{l c}   
\hline  \hline \vspace*{-2mm}&\\
Number of lanes, $n$   & $3$  \\
Number of vehicles, $m$   & $9$  \\
Maximum initial vehicle spread , $d_{long}$   & $200$ $m$ \\ 
Minimum inter-vehicle distance, $d_{\triangle}$  & $25$ $m$ \\ 
\vspace*{1mm}
Rear vehicles initial speed range, $[v^{rear}_{min}, v^{rear}_{max}]$ & $[15,25]$ $m/s$ \\
\vspace*{1mm}
Front vehicles initial speed range, $[v^{front}_{min}, v^{front}_{max}]$ & $[10,12]$ $m/s$ \\
\vspace*{1mm}
Initial speed range for ego vehicle, $[v^{ego}_{min}, v^{ego}_{max}]$ & $[10,15]$ $m/s$ \\
\vspace*{1mm}
Desired speed range for other vehicles, $[v^{d}_{min}, v^{d}_{max}]$ & $[18,26]$ $m/s$ \\
Desired speed for ego vehicle, $v^{d}_{ego}$ & $ 25 $ $m/s$ \\
Episode length, $d_{max}$ & $1000$ $m$ \\
\hline                                              
\end{tabular}
\label{table:Highway}                                
\end{table}

The initial states for each vehicle will be the inputs of the vehicle model, which is mentioned in section \ref{subsection:vehModel_ctrlModel}.

\subsection{Vehicle and Steering Control Model} \label{subsection:vehModel_ctrlModel}

Nonlinear kinematic bicycle model is used for the simulation of dynamics of the ego and surrounding vehicles. The control inputs for the kinematic bicycle model are the front steering angle $\delta_f$ and the acceleration $a$. The Intelligent-Driver Model (IDM) \cite{treiber2000congested} and a two-point visual control model of steering \cite{Two-Point} are used for the calculations of the $a$ and $\delta_f$ respectively.

\begin{figure}[H]
    \centering
    \includegraphics[width=0.8\columnwidth]{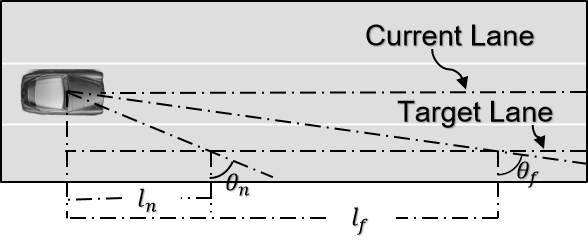}
    \caption{Two-Point Visual Control Model of Steering}
    \label{fig:Two-Point}
\end{figure}

Two-point visual control model \cite{Two-Point} is a steering control method that uses the tangent angle of two key points in near and far regions to calculate steering angle $\delta_f$ which is defined in Eq. (\ref{Two-point}).
\begin{align}
\delta_f & = k_f\theta_f + k_n\theta_n + k_I \int_{}^{} \theta_n dt \label{Two-point}
\end{align}where $\theta_n$, $\theta_f$ are near and far point angles with respect to the center of the vehicle, as shown in Fig. \ref{fig:Two-Point} and $k_f, k_n$ and $k_I$ are the tuning parameter of PI controller. Parameters which are used in the two-point visual control model are given in Table \ref{table:Two-Point}. 
Longitudinal positions of the near and far points are defined as floating points in front of the vehicle with fixed distances $l_{n}$ and $l_{f}$ for empty target lane. If there is no lane change, lateral positions of near and far points will be the center of the current lane which results with zero error for $\theta_n$ and  $\theta_f$ angles. For the lane change case, lateral positions will be the center of the target lane, as shown in Fig. \ref{fig:Two-Point}. For the occupied target lane, $l_{n}$ will remain fixed, but $l_{f}$  will be the distance between lead and current vehicles.
\begin{table}[H]
    \caption{Two-Point Visual Control Model of Steering Parameters} 
    \vspace*{-2mm}
    \centering    
    \begin{tabular}{l c}       
        \hline  \hline \vspace*{-2mm}&\\
        Distance to near point, $l_n$ &5 m \\
        Distance to far point, $l_f$ &100 m \\
        Proportional gain far  point, $k_f$ &$ 20$ \\
        Proportional gain near point, $k_n$ & $9$ \\
        Integral gain near point, $k_I$ & $10$ $s^{-1}$\\
        \hline                                              
    \end{tabular}
    \label{table:Two-Point}                                
\end{table}
Input for the two-point visual control model of steering is the target lane positions and MOBIL \cite{kesting2007general} algorithm is used to determine reference target lanes.



\subsection{Observation states and action spaces} \label{subsection:observation_actions_space}
The environment is fully observable by the ego vehicle. The ego vehicle's observing states are defined in a way to adapt to the different number of surrounding cars as shown in Table \ref{table:Observation_vector_ego} \cite{hoel2018automated}.
\begin{table}[H]
    \caption{Observation states from ego vehicle perspective} 
    \vspace*{-2mm}
    \centering    
    \begin{tabular}{l c}       
        \hline  \hline \vspace*{-2mm}&\\
        $s_1$, & Normalized ego vehicle speed $v_{ego}/v_{ego}^{d}$ \\
        $s_2$, & ego vehicle $ \begin{cases}
        1, &  \text{if there is a lane to the leftt} \\
        0, & \text{otherwise}
        \end{cases} $ \\
        $s_3$, & ego vehicle $ \begin{cases}
        1, & \text{if there is a lane to the right  } \\
        0, & \text{otherwise}
        \end{cases} $ \\
        $s_{3i+1}$, & Normalized relative position of vehicle $i$, $\Delta s_i/\Delta s_{max}$ \\
        $s_{3i+2}$, & Normalized relative velocity of vehicle $i$, $\Delta v_i/v_{max}$ \\
        $s_{3i+3}$, & $ \begin{cases}
        -1, & \text{\text{if vehicle} i \text{is two lanes to the right of ego vehicle}} \\
        -0.5, & \text{\text{if vehicle} i \text{is one lanes to the right of ego vehicle}} \\
        0, & \text{\text{if vehicle} i \text{is in the same lane as the ego vehicle}} \\
        0.5, & \text{\text{if vehicle} i \text{is one lanes to the left of ego vehicle}} \\
        1, & \text{\text{if vehicle} i \text{is two lanes to the left of ego vehicle}} 
        \end{cases} $ \\
        \hline                                              
    \end{tabular}
    \label{table:Observation_vector_ego}                                
\end{table}
\noindent where $v_{ego}^{d}$ is the maximum allowable speed for the ego vehicle, $\Delta s_{max}$ is the maximum relative position between vehicle $i$ and the ego vehicle and $v_{max}$ is the maximum allowable speed for all vehicles. The action space for the ego vehicle is depicted in Table \ref{table:action_space_ego}.
\begin{table}[H]
    \caption{Action space for the RL agent (ego vehicle)} 
    \vspace*{-2mm}
    \centering    
    \begin{tabular}{l c}       
        \hline  \hline \vspace*{-2mm}&\\
        $a_1$, & No lane change (keep current lane) \\
        $a_2$, & Lane change to the left \\
        $a_3$, & Lane change to the right \\
        \hline                                              
    \end{tabular}
    \label{table:action_space_ego}                                
\end{table}
Once the RL agent makes lane change decisions, a low-level controller as in section \ref{subsection:vehModel_ctrlModel} takes the vehicle from the current lane to the target lane.

\subsection{Intelligent-Driver Model (IDM)}
Intelligent Driver Model (IDM) \cite{treiber2000congested} is used as adaptive cruise control (ACC) functioning as a continuous-time car-following model that generates the required acceleration for the vehicle in a single lane of urban and highway traffic. Longitudinal dynamics of surrounding vehicles are simulated using IDM, as shown in Eq. (\ref{IDM1}) and (\ref{IDM2}).  
\begin{align}
\frac{dv}{dt}  = a & =   a_{max}\left(1-\left(\frac{v}{v_d }\right)^\delta - \left(\frac{d^\star(v,\Delta v)}{d }\right)^2\right) \label{IDM1}\\
& d^\star(v,\Delta v)     =  d_0 + vT +\frac{v\Delta v }{2\sqrt{ba_{max}}} \label{IDM2}
\end{align}where $v$ and $v_d$ are the current speed, and the desired speed of the vehicle, $\Delta v$ and $d$ are the speed difference and the gap between the leading vehicle and the respective vehicle. Parameters which are given in Table \ref{table:IDM} are used for tuning the simulation.   
\begin{table}[H]
\caption{Intelligent-Driver Model(IDM) Parameters} 
\vspace*{-2mm}
\centering    
\begin{tabular}{l c}   
\hline  \hline \vspace*{-2mm}&\\
Maximum acceleration, $a_{max}$  & $0.7$ $m/s^2$ \\ 
Minimum deceleration, $a_{min}$  & $-20$ $m/s^2$ \\ 
Acceleration exponent, $\delta $ & $4$\\ 
Minimum gap, $d_0$ & $2$ $m$\\
Safe time headway, $T$ & $1.6$ $s$\\
Desired deceleration, $b$ & $1.7$ $m/s^2$   \\ 
Maximum gap for empty lane, $d_{max}$ & 10000 m \\
\hline                                              
\end{tabular}
\label{table:IDM}                                
\end{table}

Longitudinal dynamics of vehicles at each lane is calculated separately according to their lane id since IDM works for single lane dynamics. If there is no vehicle in front, $\Delta v$ and $d$ is chosen as $0$ and $d_{max}$ respectively. Even though maximum accelerations of vehicles are limited by $a_{max}$, there is no limit for minimum deceleration in original IDM. We added a condition to IDM to limit minimum deceleration, as shown in Eq. (\ref{IDMLimit}).       
\begin{align}
a= 
\begin{cases}
a,& a\geq a_{min}\\
a_{min},              & \text{otherwise}
\end{cases}  \label{IDMLimit} 
\end{align}

The values of maximum acceleration corresponds to a free-road acceleration from $v = 0$ to $v = 100 \, km/h$ within $45 \, s$ and the desired or comfortable deceleration is chosen as $1.7 \, m/s^2$ \cite{treiber2000congested}.

\subsection{Minimizing Overall Braking Induced by Lane Changes (MOBIL)}

\begin{figure}[H]
    \centering
    \includegraphics[width=1.0\columnwidth]{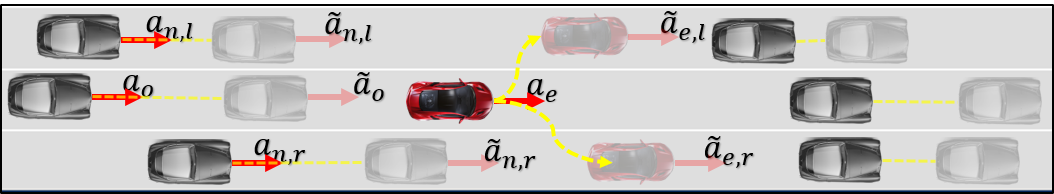}
    \caption{MOBIL algorithm decision layout }
    \label{fig:3}
\end{figure}

The MOBIL model is used to determine the safety of the lane change decisions by considering relative accelerations of the surrounding vehicles with respect to the ego vehicle. As illustrated in Fig. \ref{fig:3}, this algorithm propagates the vehicles' dynamics for a few steps further to make the appropriate decisions. The decision process of MOBIL has two steps. Firstly, it ensures that when lane change occurs, the new follower vehicle will not decelerate too much by looking at safety criteria, which is defined in Eq. (\ref{safety_1}). \begin{align}
\tilde{a}_n > b_{safe}   \label{safety_1}
\end{align} where $\tilde{a}_n$ refers to the new acceleration of the follower after making a lane change and $b_{safe}$ is the maximum safe deceleration. Secondly, if the first safety criterion is met, MOBIL checks for the second criterion, which is a collection of acceleration gains of surrounding vehicles as in Eq. (\ref{safety_2}).  
\begin{align}
\tilde{a}_e -a_e & + p (\tilde{a}_n - a_n) + q (\tilde{a}_o - a_o)  >  a_{th} \label{safety_2}
\end{align} where $\tilde{a}_e$, $\tilde{a}_n$ and $ \tilde{a}_o$ stands for new accelerations, which are determined by using IDM for one time step, for the lane changing, new follower and old follower vehicles, respectively. $a_e$, $a_n$ and $a_o$ refer to the current accelerations for the same vehicles. $p$ and $q$ are the politeness factor for the side and rear vehicles. $a_{th}$ is the lane change decision threshold. All parameters for MOBIL are given in Table \ref{table:MOBIL}.    

\begin{table}[H]
    \caption{MOBIL Parameters} 
    \vspace*{-2mm}
    \centering    
    \begin{tabular}{l c}       
        \hline  \hline \vspace*{-2mm}&\\
        Maximum safe deceleration, $b_{safe}$ & $4$ $m^2$ \\
        Politeness factor for side vehicles, $p$ & 1 \\
        Politeness factor for rear vehicles, $q$ & 0.5 \\
        Changing threshold, $a_{th}$ & 0.1 m/s2\\
        \hline                                              
    \end{tabular}
    \label{table:MOBIL}                                
\end{table}
\vspace*{-3mm}


The politeness factors ($p$ and $q$) depicted in Table \ref{table:MOBIL} are chosen in terms of safety and performance criteria based on heuristics, by which the MOBIL algorithm has achieved its best performance. Moreover, in this work, USA traffic rules are applied, which allow both right and left overtake.

\section{Hyper-parameters} \label{section:results}
There is a considerable number of hyper-parameters that can be optimized, among which we have focused on neural network (NN) architecture and reward function that has the highest impact on the overall performance.

\subsection{Neural network architecture}

Following some trial and errors and a simple grid search optimization, we came up with a simple dense multi-layer perceptron (MLP) with $\{64,128,128,64\}$ as the number of neurons in four hidden layers all activated with $tanh$ as the activation function. Keeping the network simple, not only accelerates the training time, but it also prevents overfitting. The NN receives an observation vector and produces an estimated $Q-$value for each action.

\subsection{Reward function:}

In our work, the objective was to reach the maximum allowable speed in traffic while not violating the safety criteria defined in the environment and reducing the number of lane changes. Thus, the reward for encouraging the agent to speed up is defined as below:
\[
    r(s,a,s^\prime)=\left\{
                \begin{array}{ll}
                  \text{ego speed:   }(v_{current} - v_{initial})/v_{d}\\
                  \text{lane change penalty:  } -1\\
                  \text{unnecessary action mask:   } -20\\
                  \text{collision:   } -50\\
                  \text{TTC violence:   } -5\\
                  \text{goal:   } +50
                \end{array}
              \right.
  \]
  

\noindent where $v_{d}$ is the maximum allowable speed in the highway. Also, we mask out the unnecessary actions where lead to out-of-road areas to speed up the training. Besides, the safe region for time to collision (TTC) is defined as $1.8$ seconds.

\section{results}
After setting up everything that we discussed, we trained an RL agent named agent \RomanNumeralCaps{1} for $500'000$ time steps using the DQN algorithm (section \ref{subsection:DQN}) and compared the results with the reference model (MOBIL). To replicate the agents introduced in previous works that we discussed earlier, we trained another RL agent named agent \RomanNumeralCaps{2} in a static environment where other actors were not making any complex decisions, e.g., lane changing. However, to have a fair comparison, we tested all of the agents in similar settings where all of the actors incorporate MOBIL and IDM systems to perform rational decisions to drive safer and faster. Also, we trained and tested the agents in three categories of observation noises; noise-free, mid-level noise ($\%5$), and high-level noise ($\%15$). The percentage noise is the standard deviation of the observed states from the ground truth position, velocity, and acceleration with Gaussian distribution.

\begin{figure}
    \centering
    \includegraphics[width=1.0\columnwidth]{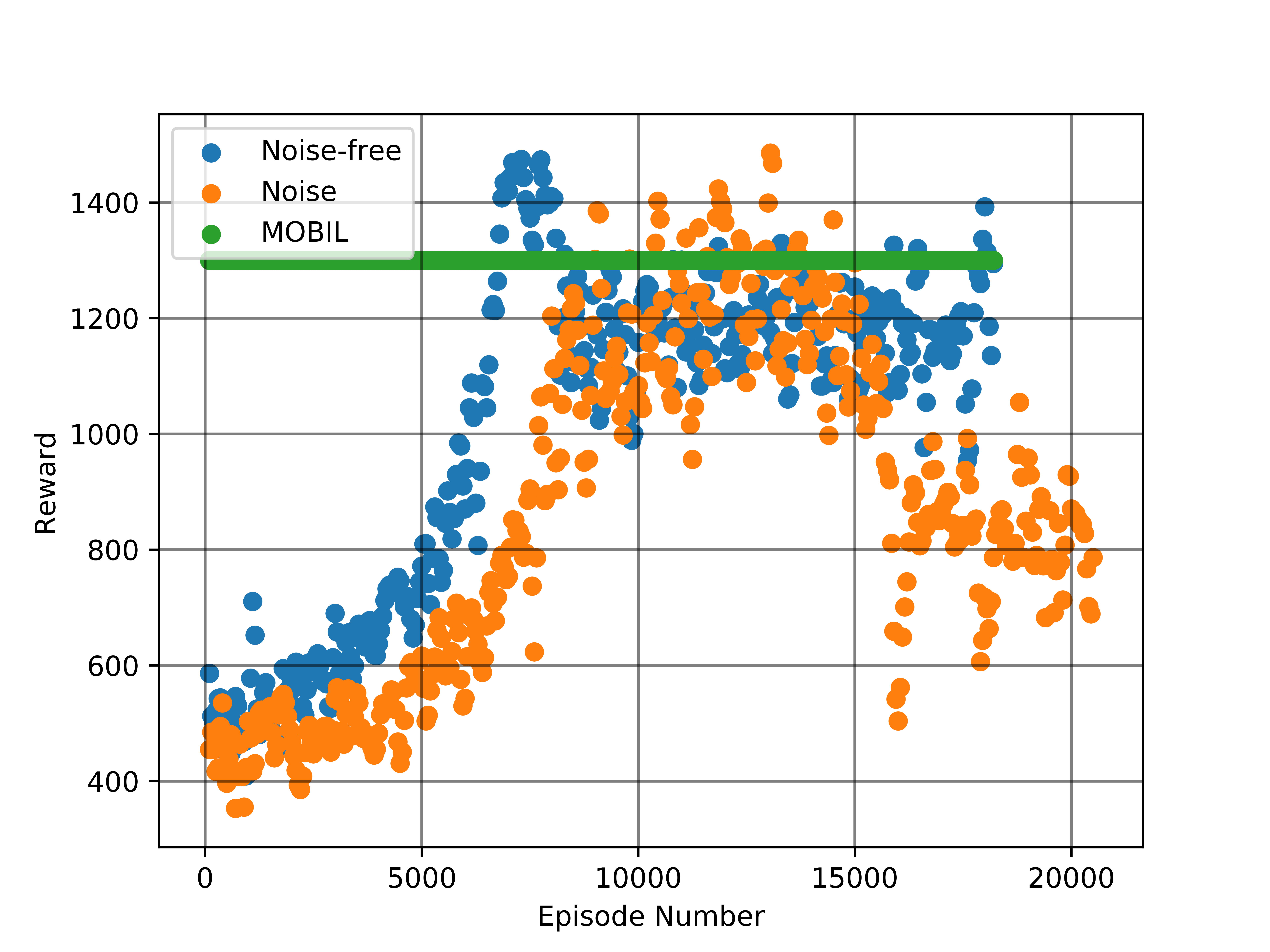}
    \caption{Performance Comparison of MOBIL algorithm and the RL agent \RomanNumeralCaps{2} in two different environments; deterministic and stochastic}
    \label{fig:training agents}
\end{figure}

The obtained mean 100-episode rewards in the training phase for the agent \RomanNumeralCaps{2} along with the MOBIL performance for (mid-level) noisy environment are illustrated in Fig. \ref{fig:training agents}. As expected, the observation noise degraded the overall performance of the RL agent during the training represented by the value of achieved rewards. Notice that, the same noisy scenarios seen by the RL agent is used to evaluate the performance of the MOBIL algorithm.

To validate the generalization of the devised RL agents and fairly compare them with a baseline which is MOBIL algorithm here, we set up a benchmark with various initial states. All of the driving vehicles are fully autonomous actors that possess car-following (IDM) and lane-changing (MOBIL) algorithms. To have a fair comparison, all of the agents are evaluated in several identical traffic scenarios, and the results are depicted in Tables \ref{table:results I} and \ref{table:results II}. 

\begin{table}[b!]
	\caption{Agent \RomanNumeralCaps{1} performance; trained on static environment with mid-level uncertainty} 
	\vspace*{-2mm}
	\centering    
	\begin{tabular}{l c c c}   
	    & Normalized &  mean reward \\
	  Observations  & (\% MOBIL) &  ($\pm$std)\\
	  \hline \hline \vspace*{-2mm}&\\
	  Noise Free & $99\%$ & $1301\pm88$\\
	  Mid-level Noise ($\pm 5\%$) & $105\%$ & $1272\pm116$\\
	  High-level Noise ($\pm 15\%$) & $120\%$ & $1006\pm205$\\
	\end{tabular}
	\label{table:results I}
\end{table}

\begin{table}[b!]
	\caption{Agent \RomanNumeralCaps{2} performance; trained on dynamic environment with mid-level uncertainty} 
	\vspace*{-2mm}
	\centering    
	\begin{tabular}{l c c c}   
	    & Normalized &  mean reward \\
	  Observations  & (\% MOBIL) &  ($\pm$std)\\
	  \hline \hline \vspace*{-2mm}&\\
	  Noise Free & $87\%$ & $1135\pm105$\\
	  Mid-level Noise ($\pm 5\%$) & $75\%$ & $988\pm176$\\
	  High-level Noise ($\pm 15\%$) & $58\%$ & $701 \pm 243$\\
	\end{tabular}
	\label{table:results II}         
\end{table}

As noticed in Table \ref{table:results I}, the trained agent \RomanNumeralCaps{1} could capture the underlying dynamics of the highway traffic and the behavior of surrounding actors with uncertain sensory data. The performance of the agent is validated in several test cases, and the results are depicted in comparison with the MOBIL algorithm as a baseline. The results are significantly superior to the rule-based MOBIL algorithm, though the agent is trained on the mid-level stochastic environment, it performs considerably well on the more uncertain environment which infers its robustness to the uncertainty and its applicability to the real-world utilization.

\begin{figure}
    \centering
    \includegraphics[width=1.0\columnwidth]{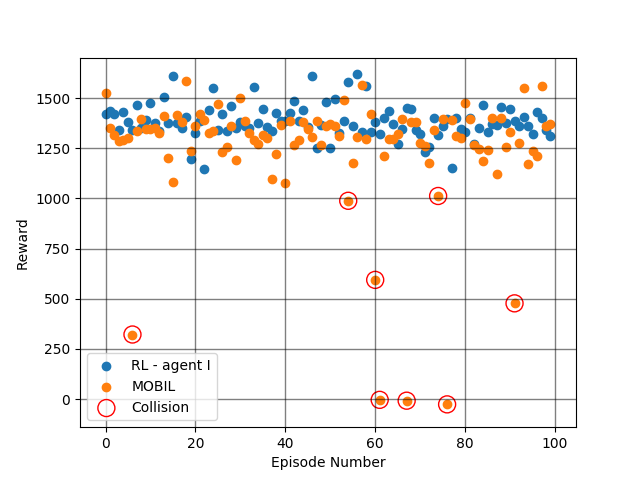}
    \caption{Performance Comparison of MOBIL and RL agent \RomanNumeralCaps{1} in 100 episodes with mid-level uncertainty}
    \label{fig:test agents}
\end{figure}

\begin{figure}
    \centering
    \includegraphics[width=1.0\columnwidth]{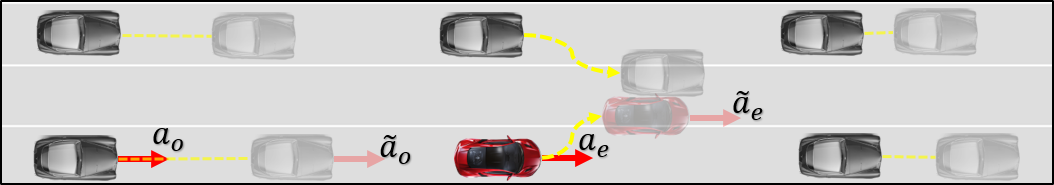}
    \caption{Weakness of MOBIL behavior in highway traffic environment}
    \label{fig:mobil}
\end{figure}

To take a closer look at the performance of the agent \RomanNumeralCaps{1} compared to MOBIL algorithm, both are evaluated for $100$ episodes, where the achieved rewards for each episode is explicitly shown in Fig. \ref{fig:test agents}. The performed RL agent \RomanNumeralCaps{1} in Fig. \ref{fig:test agents} has become so robust to the uncertain input states, but MOBIL could not recover from the uncertainty applied to the environment, and its performance has degraded dramatically. In $100$ episodes, the mean reward of RL agent \RomanNumeralCaps{1} was comparable to MOBIL, where MOBIL resulted in $8$ collisions but the RL agent did not make any collisions which are desirable. As shown in Fig. \ref{fig:mobil}, when multiple agents have MOBIL governing their behavior, at an instance such collision might happen for ego MOBIL while RL agents can learn to recover from such scenarios.

The agent \RomanNumeralCaps{2} is well trained on a static environment (see Fig. \ref{fig:training agents}). To the best of authors' knowledge, this type of environment and training were what most contender works in the literature used \cite{mukadamtactical, hoel2018automated}. By comparing the results of agent \RomanNumeralCaps{2} in Table \ref{table:results II} with agent \RomanNumeralCaps{1}, we can conclude that training an RL agent in such static environment cannot capture the underlying dynamics and uncertainty of the real-world applications where surrounding vehicles make rational decisions such as lane changing.

Although MOBIL performs better than both RL agents for the noise-free case, agent \RomanNumeralCaps{1} has shown better performance for the noisy environments compared to both MOBIL and agent \RomanNumeralCaps{2} as it is trained in a stochastic and dynamic environment. Furthermore, it is trivial that as the agent \RomanNumeralCaps{2} is trained in a static environment, it presets more reduced performance compared to both other agents.

Results mentioned above prove that training an RL agent in a noisy environment (close to the real-world cases) with dynamic actors driving in the scene, results in more robust and reliable performance.

\section{CONCLUSIONS}
Autonomous lane changing is a principal player for autonomous driving of levels 2 and above. In this work, we have trained a deep RL agent to perform the high-level decision making for an autonomous vehicle on the uncertain highway driving task. We have shown the superior capability of the RL in dealing with uncertain and stochastic environments compared with the rule-based baseline methods. We showed that using deep RL, the safety and agility of the ego vehicle can be balanced on-the-go, which indicates an adaptive behavior. Thus upon the demand, the performance of the agent will change with no handcrafting, aiming to achieve the desired goals. For the future work, it is planned to create a collaborative multi-agent environment where each agent cooperates to achieve a shared goal, e.g., increasing traffic flow or reducing time-of-arrival for each agent.

\addtolength{\textheight}{-12cm}   








\bibliography{references.bib}
\bibliographystyle{ieeetr}


\end{document}